\documentclass[10pt,twocolumn,letterpaper]{article}

\usepackage{iccv}
\usepackage{times}
\usepackage{epsfig}
\usepackage{graphicx}
\usepackage{amsmath}
\usepackage{amssymb}
\usepackage[table,xcdraw,dvipsnames]{xcolor}
\usepackage{enumitem}
\usepackage{multirow}
\usepackage{tabularx}
\usepackage{booktabs} %

\usepackage[breaklinks=true,bookmarks=false]{hyperref}

\colorlet{colorFst}{Green!25}       %
\colorlet{colorSnd}{SpringGreen!45} %
\colorlet{colorTrd}{Yellow!30}      %

\newcommand{\fs}{\cellcolor{colorFst}\bf}   %
\newcommand{\nd}{\cellcolor{colorSnd}}      %
\newcommand{\rd}{\cellcolor{colorTrd}}      %

\newcommand{\tabcapspace}{5pt}

\iccvfinalcopy %

\def\iccvPaperID{7} %

\ificcvfinal\fi

\usepackage{soul}

\newcommand{\boldparagraph}[1]{\vspace{0.4em}\noindent{\bf #1}}

\newcommand{\Ours}{APNet}
\newcommand{\OursFusion}{GAF}

\begin{document}

\title{{APNet:} Urban-level Scene Segmentation of Aerial Images and Point Clouds}

\author{Weijie Wei \hspace{2.5em} Martin R. Oswald \hspace{2.5em} Fatemeh Karimi Nejadasl \hspace{2.5em} Theo Gevers\\
University of Amsterdam\\
{\tt\small \{w.wei2, m.r.oswald, f.kariminejadasl, th.gevers\}@uva.nl}
}

\maketitle
\ificcvfinal\thispagestyle{empty}\fi

\begin{abstract}
In this paper, we focus on semantic segmentation method for point clouds of urban scenes. Our fundamental concept revolves around the collaborative utilization of diverse scene representations to benefit from different context information and network architectures. To this end, the proposed network architecture, called APNet, is split into two branches: a point cloud branch and an aerial image branch which input is generated from a point cloud. To leverage the different properties of each branch, we employ a geometry-aware fusion module that is learned to combine the results of each branch. Additional separate losses for each branch avoid that one branch dominates the results, ensure the best performance for each branch individually and explicitly define the input domain of the fusion network assuring it only performs data fusion. Our experiments demonstrate that the fusion output consistently outperforms the individual network branches and that APNet achieves state-of-the-art performance of 65.2 mIoU on the SensatUrban dataset. Upon acceptance, the source code will be made accessible.

\end{abstract}

\section{Introduction} \label{sec:introduction}

\begin{figure}[thb] \centering
    \makebox[0.23\textwidth]{
    \includegraphics[width=0.23\textwidth]{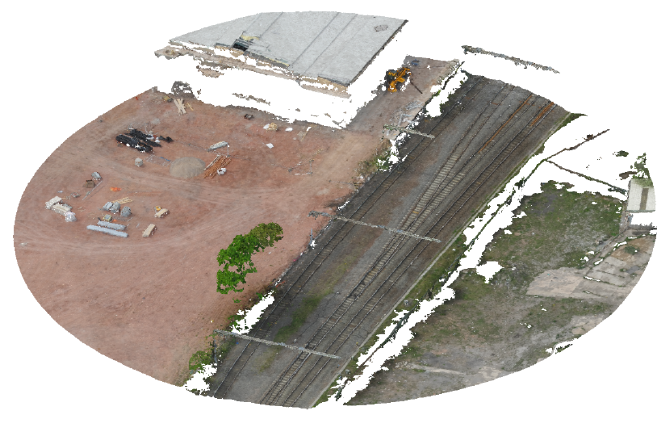}
    }
    \\
        \makebox[0.162\textwidth]{\footnotesize a) Colored point cloud}
    \\
    \includegraphics[width=0.23\textwidth]{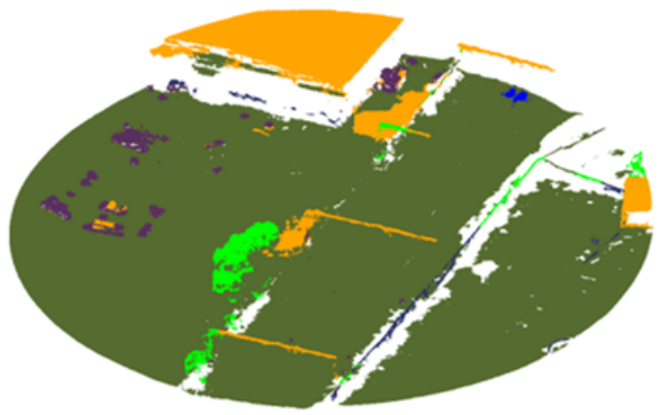}
    \includegraphics[width=0.23\textwidth]{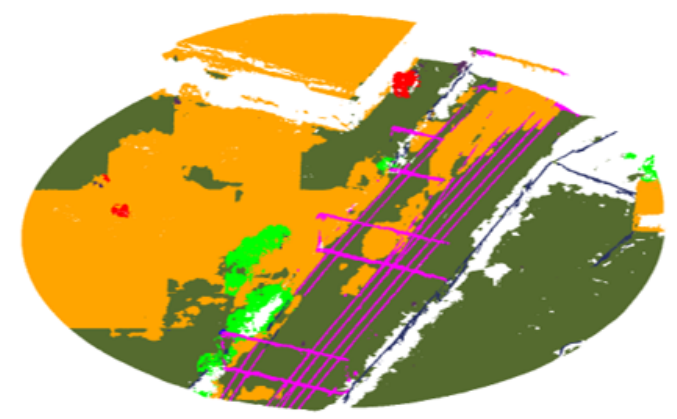} 
    \\
        \makebox[0.2\textwidth]{\footnotesize b) Prediction of P-branch}
        \makebox[0.2\textwidth]{\footnotesize c) Prediction of A-branch}
    \\
    \includegraphics[width=0.23\textwidth]{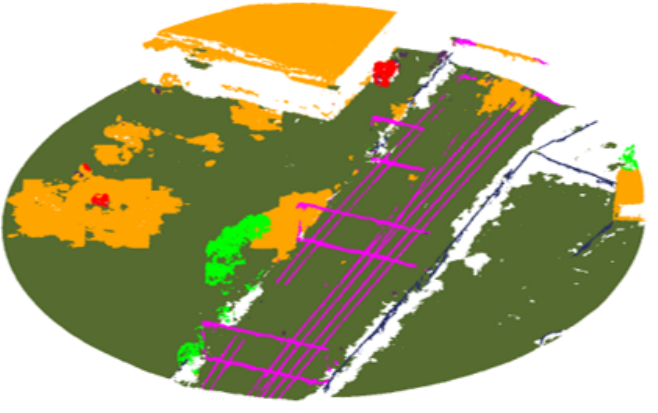}
    \includegraphics[width=0.23\textwidth]{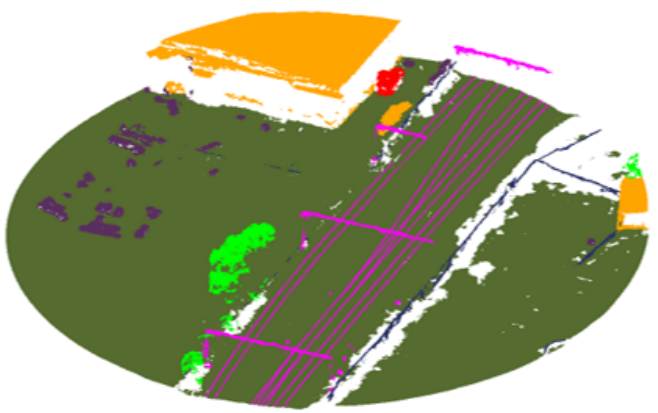} 
    \\
        \makebox[0.2\textwidth]{\footnotesize d) Prediction of \Ours{}}
        \makebox[0.2\textwidth]{\footnotesize e) Ground Truth}\\[4pt]
    \caption{\textbf{\Ours{} Segmentation.}
    Starting from a colored input point cloud (a) the data is fed into two separate branches: a point-cloud branch (b) and an aerial image branch (c).
    The key idea is to exploit the advantages of both branches regarding spatial context and network architectures.
    The results of both branches is then merged with a fusion network.
    \Ours{} achieves a better result, which is much closer to the ground truth than the solution of individual branches.    
    } 
    \label{fig:teaser}
    \vspace{-3mm}
\end{figure}

Urban-level point cloud segmentation is an important stepping stone for semantic scene understanding for various applications like autonomous driving, robotics, large-scale map creation or mixed reality~\cite{li_deep_2020, guo_survey_pointcloud_2020}. The majority of urban semantic segmentation methods can be categorized to either use aerial / birds-eye-view image data~\cite{yang2017convolutional, zhao2018classifying} or 3D point cloud data~\cite{pointnet2, PointCapsule, randlanet_2020}.

On the one hand, 2D/2.5D image-based approaches benefit from the simple data structure that allows for highly effective aggregation of large spatial contexts which is useful for semantic inference and for which a large pool of network architectures exist~\cite{lang2019pointpillars, milioto_rangenet_2019, yang2017convolutional, zhao2018classifying}.
However, these methods are limited to resolve full 3D shapes and spatial context along the gravity directions. 

On the other hand, point cloud-based approaches can leverage full 3D spatial context, but context aggregation and high detail levels are generally much more expensive to progress and are thus more limited in spatial resolution and context reasoning.
Unlike images, as they may suffer from large color variations due to changing weather conditions or day-to-night cycles, point clouds are more robust to these phenomena~\cite{li_deep_2020}.
However, point clouds are more challenging to process due to their irregular and non-uniform structure.
Similarly, many established network architectures exist for point cloud processing~\cite{pointnet, pointnet2, thomas_kpconv_2019, PointCapsule,randlanet_2020}.

We argue that semantic reasoning in both domains has advantages and disadvantages, \eg incorporating a larger context within a 2D domain enhances the recognition capabilities of flat and large objects, whereas small objects with a 3D spatial extension are more effectively detectable within the 3D domain. With this objective in mind, our primary aim is to leverage and combine the best properties of both domains to propose a unified semantic segmentation approach that synergistically learns from both.

Recent papers show impressive results on vehicle-based point cloud datasets by combining different representations~\cite{liong_amvnet_2020, tang_searching_2020, xu_rpvnet_2021}. However, their corresponding representations, \eg range-view and voxelization, are less suited for UAV-based datasets. To address the aforementioned objectives, we propose \Ours{}, which concurrently operates within the aerial image domain and the point cloud domain. Exemplary results of \Ours{} are depicted in Fig.~\ref{fig:teaser}.
Our \textbf{contributions} can be summarized as follows:
\begin{itemize}
    \item We introduce \Ours{}, an effective network architecture for urban-level point cloud segmentation that leverages differences in domain properties regarding network architectures and spatial context by following a multi-branch where each branch is specialized for a particular domain.
    \item We propose a geometry-aware fusion module that introduces the geometric information of the original points into the process of feature fusion of two branches and achieve a better performance. 
    \item Our experiments demonstrate the efficacy of \Ours{} by attaining state-of-the-art performance on the SensatUrban dataset~\cite{sensaturban_cvpr}.
\end{itemize}

\section{Related Work} \label{sec:relatedwork}

\subsection{Single Representation for Point Cloud Segmentation}

In recent years, various deep learning-based methods are proposed for point cloud segmentation. These methods can be grouped into three categories based on their representation: projection-based, voxelization-based and point-based methods.
The aim of both the projection-based and voxelization-based methods is to transform 3D point clouds to a regular representation and then use off-the-shelf networks to extract the features. In contrast, point-based methods directly process irregular point clouds.

\boldparagraph{Projection-based representation.}
Deep learning has made great strides in 2D computer vision tasks, leading researchers to apply the well-established 2D networks to 3D tasks.
Lawin et al.~\cite{lawin2017deep} propose a 3D-2D-3D pipeline to solve point cloud segmentation. 
They project a point cloud onto multi-view 2D planes and feed the resulting images to a 2D segmentation network.
The final semantic per-point label is obtained by fusing the pixel-level predictions.
Although the multi-view strategy can alleviate occlusion, the pre- and postprocessing are inefficient and the results are sensitive to viewpoint selection.
Furthermore, multi-view projection is typically used for a single scene or object, whereas urban-scale point clouds usually result in more occlusion. 
Other approaches utilize range-view planes as an intermediate representation for point cloud datasets collected by rotating laser scanner~\cite{squeezeseg, squeezesegv2, milioto_rangenet_2019}, which is a typical sensor for autonomous vehicles.
In this scenery, the egocentric spherical representation can retain more information in contrast to a single plane representation.
However, this representation is not well-suited for UAV-based datasets as it results in server occlusion due to the inconsistency of laser direction and projection direction.
Inspired by these methods, we propose to project the point cloud onto aerial-view plane that is perpendicular to the laser. The one-time aerial-view projection is efficient and avoids information loss caused by occlusion as much as possible.

\boldparagraph{Voxelization-based representation.}
These methods convert a point cloud into a discrete representation, such as cubic voxels, and then use a 3D convolution neural network (CNN) to compute the features~\cite{voxelnet, segcloud}. 
This representation naturally preserves the neighborhood structure of 3D point clouds but 3D CNNs are memory and computation-intensive.
These costs increase dramatically in outdoor scenarios due to the sparsity of points leading many empty voxels.
Although some methods use sparse convolution to reduce these costs, the discretization unit is non-trivial to determine~\cite{tang_searching_2020, SparseConvNet}.
Furthermore, urban-level datasets often contain heterogeneous objects, ranging from tiny bikes and to huge buildings and streets, which makes them unsuitable for voxelization-based methods.

\boldparagraph{Point-based representation.}
Point-based methods directly process irregular point clouds by different means, \eg multi-layer perceptron, point convolution or graph-based operations. 
MLP-based networks usually stack multiple MLPs with a feature aggregation module in accordance to the convolution layers with a subsequent pooling layer in 2D neural network~\cite{pointnet, pointnet2, randlanet_2020}. 
Furthermore, point convolution simulates powerful 2D convolution in 3D space by utilizing a parametric continuous convolution layer~\cite{wang_pccn} or a group of kernel points as reference points~\cite{thomas_kpconv_2019}.
Point-based methods are applicable to various datasets because they do not rely on transforming a point cloud to other intermediate representations.
So far, there are only point-based methods proposed for urban-level point cloud segmentation.
For instance, both EyeNet~\cite{Yoo_2023_CVPRW} and LGS-Net~\cite{LGSNet} utilize a point-based network, namely RandLA-Net~\cite{randlanet_2020}, as their backbone.
MRNet exploits multiple 3D receptive fields and LGS-Net emphasizes the utilization of geometric information.
Du \etal~\cite{PushBoundary}, using KPConv~\cite{thomas_kpconv_2019} as the backbone, exploit a multi-task framework to achieve both boundary localization and semantic segmentation.
Huang \etal~\cite{lcpformer} improve a transformer-based network by applying a local context propagation module to ensure message passing among neighboring local regions. 
Despite numerous efforts, point-based methods remain computationally intensive. 
Increasing the receptive field of point-based methods is challenging, whereas this can be easily accomplished in highly-optimized 2D networks.

\vspace{2mm}
In summary, numerous methods have been proposed for point cloud segmentation, but a handful of them are suitable for urban-level point cloud segmentation.
Additionally, single representations have their limitations.
For urban-level point cloud segmentation, geometric information and large receptive fields are equally crucial. Therefore, we propose \Ours{} to combine aerial-views and point-based representations.
To the best of our knowledge, we are the first to propose a hybrid method to handle urban-level point cloud segmentation.

\subsection{Hybrid Representation for Point Cloud Segmentation}
There are also a number of methods that combine different representations. One common strategy is to parallelize multiple networks processing different representations and combining features at different levels.
SPVNAS~\cite{tang_searching_2020}, Cylinder3D~\cite{zhu_cylindrical_2020} and DRINet~\cite{DRINet} share the concept of paralleling voxel-point architectures. 
SPVNAS~\cite{tang_searching_2020} introduces a sparse voxel convolution and combines voxel-wise and point-wise features in different stages.
Cylinder3D~\cite{zhu_cylindrical_2020} imposes a point refinement module at the end of the network, which sums voxel-wise and point-wise features followed by three fully-connected layers.
DRINet~\cite{DRINet} introduces a voxel-point iteration module to iteratively interact between two features. 
RPVNet~\cite{xu_rpvnet_2021} consists of three branches, \ie range-view, point-wise and voxel branches.
A gated attention module generates coefficients for a linear combination that point-wisely combines information from three branches. 
These methods combine features either by a simple addition or point-wise combination but fall short to incorporate features from neighbour points.
AMVNet~\cite{liong_amvnet_2020} addresses this issue by training a small assertion-based network and feeding information from neighbours into it to generate final predictions.
However, in the small network, only semantic predictions, \ie class-wise probability scores, are considered. Hence, deeper features with richer contextual information are ignored.
In conclusion, hybrid methods leverage prior knowledge from different representations to enhance features to achieve better performance. Nevertheless, the fusion module is often naive and the information from neighbour points is ignored.

Therefore, in this paper, we propose a simple yet effective fusion module that takes both contextual and geometric features as input and exploits positional relationships among neighbour points to generate descriptive features. In contrast to previous methods, our approach effectively incorporates information from neighboring points and achieves better performance on urban-level point cloud segmentation tasks.

\section{Methodology} \label{sec:method}
\begin{figure*}
  \centering
   \includegraphics[width=\linewidth]{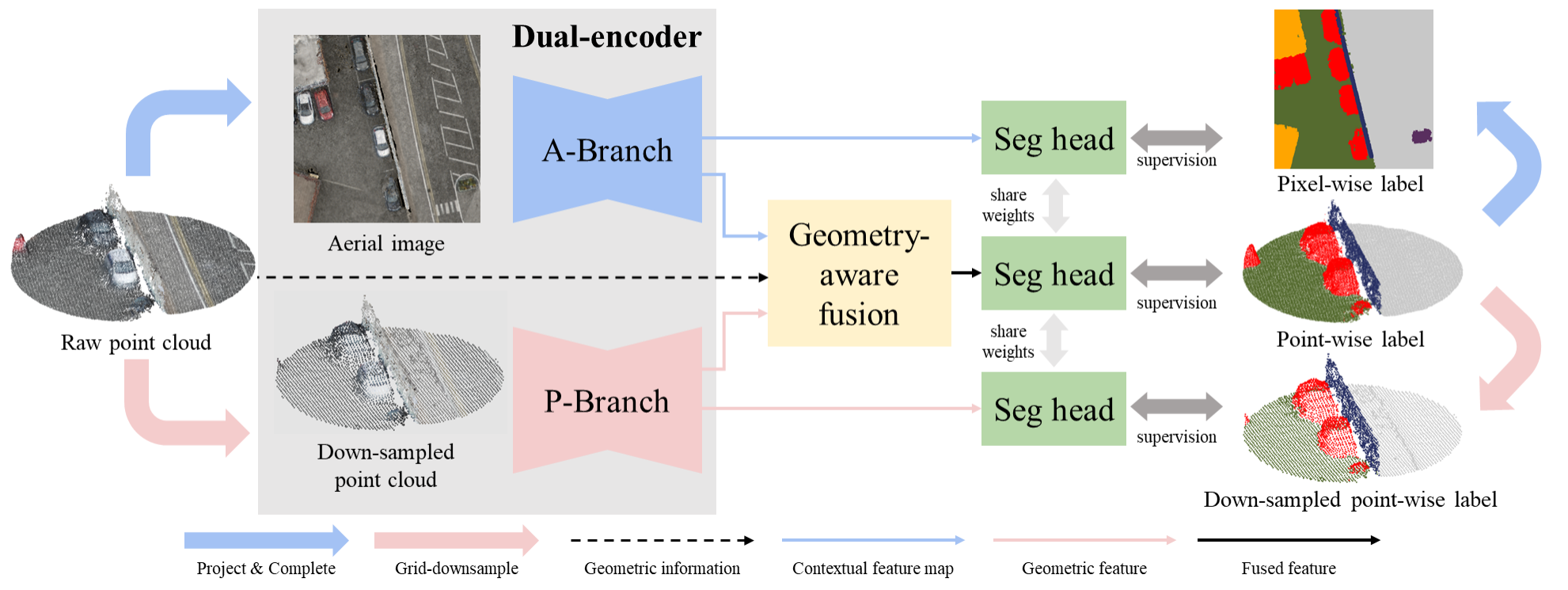}
   \caption{\textbf{Architecture overview of \Ours{}.} 
   The network consists of a dual-encoder, a geometry-aware fusion module and three segmentation heads that operate in different domains. The two representations of a sample, \ie aerial image and down-sampled point cloud, are fed into the dual-encoder. Their outputs are passed to the fusion module for feature aggregation. Finally, the features are sent to the segmentation head for point-wise segmentation. }
\label{fig:overview}
\vspace{-3mm}
\end{figure*}
In this section, we first present the problem statement.
Then, we discuss the different components of our \Ours{}, \ie the dual-encoder and the \OursFusion.
Finally, we explain the segmentation heads and define loss functions.

\boldparagraph{Problem statement.} %
Given a colored point cloud $\mathbf{P}=\{(p_k, c_k)\}_{k=1}^N$ with $N$ point coordinates $p_k=(x_k, y_k, z_k) \in \mathbb{R}^3$ and colors $c_k=(r_k, g_k, b_k)\in \mathbb{R}^3$, the aim is to compute the corresponding semantic labels $\mathbf{L}=\{(l_k)\}_{k=1}^N$ for every point.
We train a deep learning model $h(\cdot | \theta)$ with parameter $\theta$ by minimizing the difference between the prediction $\mathbf{L}=h(\mathbf{P}| \theta)$ and corresponding ground truth label set $\mathbf{\hat{L}}$. 
The urban-level point cloud datasets are obtained by UAVs. 

\subsection{Dual-encoder}
The key idea of our approach is to split up the label prediction into two different domains: an aerial (A)-branch and a point-based (P)-branch to leverage the advantages of using different spatial contexts that corresponding 2D vs. 3D network architectures have.
The output of both branches is then fused within a geometry-aware fusion (GAF) module as illustrated in Fig.~\ref{fig:overview}.
Rather than fusing the label predictions of each branch $\mathbf{L}^a$ and $\mathbf{L}^p$, the \OursFusion~operates on intermediate feature representations $\mathbf{F}^a$ and $\mathbf{F}^p$ for a more informed label decision process.
We detail both branches in the following paragraphs.

\boldparagraph{Aerial image branch.} To obtain a pseudo aerial image of a point cloud, we first project it to an aerial view by an orthographic projection. 
Assuming that the gravity direction is aligned with the z-axis, each point $p_k = (x_k, y_k, z_k)$ is converted to a pixel $p_i = (u_i, v_i)$ via a mapping $\rho: \mathbb{R}^3 \mapsto \mathbb{R}^2 $, as defined by
\begin{equation}
    (u_i, v_i)^T= \rho(p_k) = \left(\left \lfloor \frac{x_k}{s} \right \rfloor, \left \lfloor \frac{y_k}{s} \right \rfloor \right)^T \enspace,
    \label{eq:aerial_proj}
\end{equation}
where $i$ is the index of a pixel and $s$ is the quantization unit, \ie pixel size. 
By aggregating all 3D points into pixels, we obtain the initial aerial image $ \mathbf{I}^{init} \in \mathbb{R}^{H \times W \times 3} $.
Note that the mapping function $\rho$ is a many-to-one function and we only preserve the properties, \eg color and label, of the highest point in the final image. 
Moreover, due to the sparsity of LiDAR points, a pseudo image created from the projection of a point cloud must be completed because, unlike a genuine aerial image, it contains both valid and null pixels.
A pixel is considered valid if it covers a minimum of one LiDAR point and is regarded as null otherwise.
During the completion, valid pixels are dilated.
When the eight neighbour pixels of a null pixel have more than two distinct values, its value is updated by the value that occurs most frequently among its neighbouring pixels.
After the completion, we obtain the input aerial image $\mathbf{I} \in \mathbb{R}^{H \times W \times 3}$.
The same projection and completion are operated on labels.

Due to the simplicity of our method, A-branch can be any end-to-end 2D semantic segmentation network.
Its output is defined as follows:
\begin{equation}
    \mathbf{F}^a = h^a(\mathbf{I} | {\theta}^a) \enspace,
\end{equation}
where $h^a(\cdot | {\theta}^a)$ is the A-branch network and $\mathbf{F}^a \in \mathbb{R}^{H \times W \times C}$.

\boldparagraph{Point cloud branch.} The original point cloud provides precise geometric information and is of importance in the ultimate evaluation. However, the spatial distribution of a point cloud is not uniform and local points with the same semantics tend to contain homogeneous information.
To ensure the points are sampled uniformly and to increase the network's receptive field, grid-downsampling is frequently used~\cite{thomas_kpconv_2019, randlanet_2020}.
We follow KPConv~\cite{thomas_kpconv_2019} to perform grid-downsampling on the original point cloud, which creates a barycenter point for each non-empty grid, with the average values of all points within the same grid serving as the new properties of the barycenter point. 
The downsampled points are denoted as %
\begin{equation*}
  \mathbf{P}^d=\{(p_k, c_k)\}_{k=1}^{N^d} \enspace.
\end{equation*}
Similar to the flexibility of the A-branch, the P-branch can be easily replaced by any point-based network and is denoted by $h^p(\cdot | {\theta}^p)$.
By passing downsampled points to the P-branch, a point-wise feature representation is obtained:
\begin{equation}
    \mathbf{F}^p = h^p(\mathbf{P}^d | {\theta}^p) \enspace,
\end{equation}
where $\mathbf{F}^p \in \mathbb{R}^{N^d \times C}$.

For both the P-branch and A-branch, instead of using ultimate semantic predictions of two base models, we use the high-dimensional features from the intermediate layers of two base models.

\subsection{Geometry-aware Fusion Module}

In point cloud segmentation, many methods~\cite{thomas_kpconv_2019, randlanet_2020, xu_rpvnet_2021} commonly employ a preprocessing step to achieve a uniform point density. This is typically achieved through grid-downsampling, wherein the point cloud is transformed into a gird-based represetnation. During the training and validation stages, only the newly generated points are processed within the network.
The postprocessing, namely up-sampling, only occurs during the testing phase, where the labels of original points are determined based on the predictions of their nearest neighbouring points. 
However, this pipeline fails to include the features of other neighbouring points and the geometric information of the original point cloud throughout the training process.
\begin{figure}[t]
  \centering
   \includegraphics[width=0.95\linewidth]{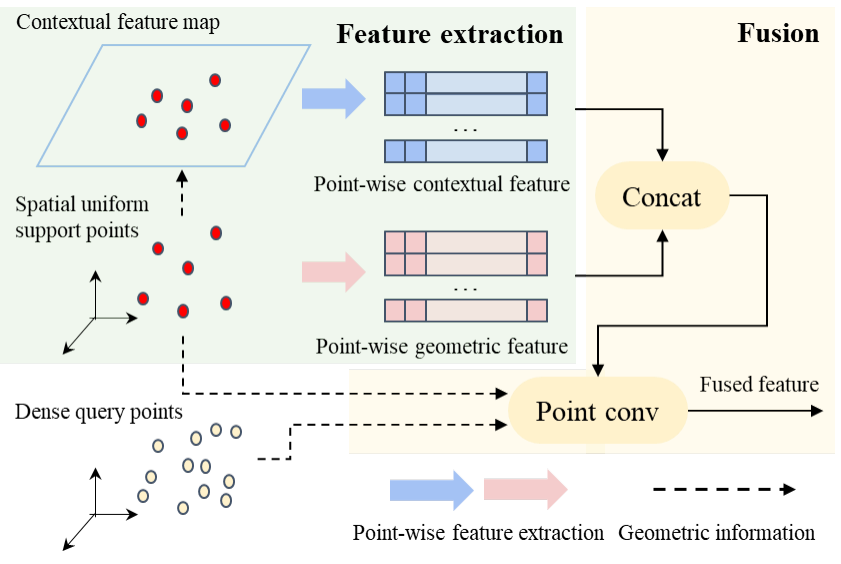}
   \caption{ \textbf{Geometry-aware fusion module} includes feature extraction and fusion. Given support points and a contextual feature map, point-wise contextual features are extracted and concatenated with point-wise geometric features. The concatenated features and geometric information of both query points and support points are fed into a point convolution to aggregate geometric context information for generating the fused output features.}
   \label{fig:fusion_module}
\vspace{-3mm}
\end{figure}
To address this, we employ a skip connection to convey geometric information of the original point cloud to the fusion module and utilize a point convolution to gather features of neighbour points.
Our \OursFusion{}~module includes two parts, namely feature extraction and fusion, as illustrated in Fig.~\ref{fig:fusion_module}.

The feature extraction is performed at the downsampled point level to reduce the computational complexity.
For a given point belonging to downsampled points $p_k^d \in \mathbf{P}^d$, its features are computed from the outputs of two branches.
Specifically, the output of the P-branch, which is in the form of a point-wise feature and thus ready to use, is denoted as $f_k^d$ for a specific point $p_k^d$.
In the other hand, for the pixel feature, unlike the quantization operation in generating the aerial image, the bilinear interpolation and the precise 2D coordinates of the point $p_k^d$, \ie $(u_k, v_k)=({x_k^d}/{s}, {y_k^d}/{s})$, %
are used to obtain pixel feature:
\begin{equation}
    f_k^a = \sum_{u,v \in \delta(k)} \phi(u_k, v_k, u, v)\mathbf{F}^a(u, v) \enspace,
\end{equation}
where $\delta(k)$ is the set of the four neighboring pixels of point $k$ and $\phi(\cdot)$ computes the bilinear weights.

The process of feature involves the concatenation of features derived from the two branches and a point convolution.
A point convolution, \eg KPConv~\cite{thomas_kpconv_2019}, is defined as follows,
\begin{equation}
    f_k=\mathcal{G}(p_k) = \sum_{p_l \in \mathcal{N}_{p_k} }g(p_k - p_l)f_l \enspace,
\end{equation}
where $\mathcal{G}$ represents the point convolution, while $g( \cdot )$ denotes the kernel function that computes the weights based on the vector from target point $p_k$ to one of its neighbouring points $p_l$. $f_l$ is the concatenated feature of point $p_l$ from feature extraction module and $\mathcal{N}_{p_k}$ refers to the neighbouring points of point $p_k$. In summary, the feature of a target point is obtained by weighted sum the features of its neighbouring points.

For each single point convolution, we use one point from a pre-defined query set $\mathbf{P}^q$ as the target point and obtain its features based on its neighbouring points from a pre-defined support set $\mathbf{P}^s$. Note that the neighbouring point set, denoted as $\mathcal{N}_{p_k}$, is a subset of the support set $\mathbf{P}^s$. This subset is generated by considering the distances between the neighbouring points and the target point $p_k$.
A common practice is to use a same point cloud, \eg a downsampled point cloud, for both the query set and support set~\cite{thomas_kpconv_2019}, which is denoted as the naive GAF module, as discussed in Sec.~\ref{sec:ablation}.
In this way, the entire network works at the level of downsampled points.
Nevertheless, our investigations indicate that the performance is negatively affected by disregarding the precise geometric information of the original points.
To address this, we opt to utilise the original points $\mathbf{P}$ instead of the downsampled points $\mathbf{P}^d$ as the query set, which implies that we set $\mathbf{P}^q = \mathbf{P}$.
The fused feature $\mathbf{f}_k^\text{fused}$ of point $p_k$ is obtained by $\mathbf{f}_k^\text{fused} = \mathcal{G}(p_k)$ and the feature set is defined as $ \mathbf{F}^\text{fuse} = \{\mathbf{f}_k^\text{fused} | k = 1,2,..N \}$.

In summary, the feature extraction operates at the level of downsampled points and the feature fusion incorporates the precise geometric information of the original points during the training stage, which enhances the accuracy.

\subsection{Segmentation Heads and Loss function}
The segmentation heads are a set of convolutional layers with $1\times1$ kernel compressing the channel from a high dimension to a low one, namely the number of categories.
The final output of the model is defined by:
\begin{equation}
    \mathbf{Pred}^\text{rep} = \mathrm{Conv}_{1\times1}^m(\mathbf{F}^\text{rep}) \enspace,
\end{equation}
where $\mathbf{Pred}^\text{rep} \in \mathbb{R}^{1\times N_\text{classes}}$ is the probabilistic prediction based on the feature $f^\text{rep}$ and $rep \in \{a, p, fused\}$ stands for aerial, point-wise or fused representation.
$\mathrm{Conv}_{1\times1}^m$ means $1\times 1$ a convolutional layer is repeated for $m$ times.

Two class-balanced loss function is used, \ie weighted cross-entropy (WCE) with inverse frequency~\cite{cortinhal2020salsanext} and Lov{\'a}sz-softmax loss~\cite{berman2018lovasz}.
The WCE loss is applied between the output of three segmentation heads and corresponding ground-truths:
\begin{equation}
    \mathcal{L}_1^\text{rep} = \mathcal{L}_\text{WCE}(\mathbf{Pred}^\text{rep}, \mathbf{\hat{L}}) \enspace,
\end{equation}
Note that although three representations share the same segmentation head and the loss function, the ground-truths $\mathbf{\hat{L}}$ are different. 
The pixel-wise label, grid-downsampled point label and the label for raw points are applied to aerial, point-wise and fused predictions respectively.
The Lov{\'a}sz-softmax loss is only applied to the fused representation:
\begin{equation}
    \mathcal{L}_2 = \mathcal{L}_\text{Lovasz} (\mathbf{Pred}^\text{rep}, \mathbf{\hat{L}}) \enspace,
\end{equation}
Eventually, the overall loss is calculated as:
\begin{equation}
    \mathcal{L}_\text{all} = \sum_{\text{rep}=\{a,p,\text{fused}\}} \alpha^\text{rep} \mathcal{L}_1^\text{rep} + \beta \mathcal{L}_2 \enspace.
\end{equation}
where $\alpha$ and $\beta$ are the factors to adjust the scale of loss functions.

\section{Experiments} \label{sec:experiments}
In this section, we introduce the implementation details of our \Ours{} in Sec.~\ref{sec:exp_setup}. Then we compare the proposed model with SOTA models on the SensatUrban dataset~\cite{sensaturban_cvpr} in Sec.~\ref{sec:sota_res}. Finally, the effectiveness of all components are analyzed in Sec.~\ref{sec:ablation}.

\newcommand{\spheading}[2][4.3em]{%
  \rotatebox{90}{\parbox{#1}{\raggedright #2}}}

\begin{table*}[t] 
  \centering    
    \resizebox{\textwidth}{!}{
        \begin{tabular}{lccccccccccccccc}
        \hline
          Method &
          OA &
          mIoU &
          \spheading{ground} & 
          \spheading{vegetation} & 
          \spheading{building} & 
          \spheading{wall} & 
          \spheading{bridge} & 
          \spheading{parking} & 
          \spheading{rail} & 
          \spheading{traffic road} & 
          \spheading{street furniture} & 
          \spheading{car} & 
          \spheading{footpath} & 
          \spheading{bike} & 
          \spheading{water} \\ 
        \hline
        PointNet~\cite{pointnet}         & 80.8 & 23.7 & 67.9 & 89.5 & 80.1 & 0.0  & 0.0  & 3.9  & 0.0  & 31.6 & 0.0  & 35.1 & 0.0  & 0.0 & 0.0  \\
        PointNet++~\cite{pointnet2}      & 84.3 & 32.9 & 72.5 & 94.2 & 84.8 & 2.7  & 2.1  & 25.8 & 0.0  & 31.5 & 11.4 & 38.8 & 7.1  & 0.0 & 56.9 \\
        TangentConv~\cite{TagentConv}    & 77.0 & 33.3 & 71.5 & 91.4 & 75.9 & 35.2 & 0.0  & 45.3 & 0.0  & 26.7 & 19.2 & 67.6 & 0.0  & 0.0 & 0.0  \\
        SPGraph~\cite{SPG}               & 85.3 & 37.3 & 69.9 & 94.6 & 88.9 & 32.8 & 12.6 & 15.8 & 15.5 & 30.6 & 22.9 & 56.4 & 0.5  & 0.0 & 44.2 \\
        SparseConv~\cite{SparseConvNet}  & 88.7 & 42.7 & 74.1 & 97.9 & 94.2 & 63.3 & 7.5  & 24.2 & 0.0  & 30.1 & 34.0 & 74.4 & 0.0  & 0.0 & 54.8 \\
        KPConv~\cite{thomas_kpconv_2019} &  93.2 & 57.6 & \fs 87.1 & \fs 98.9 & 95.3 & \rd 74.4 & 28.7 & 41.4 & 0.0  & 55.9 & \nd 54.4 & \nd 85.7 & 40.4 & 0.0 & \fs 86.3 \\
        RandLA-Net~\cite{randlanet_2020} & 89.8 & 52.7 & 80.1 & 98.1 & 91.6 & 48.9 & 40.6 & 51.6 & 0.0  & 56.7 & 33.2 & 80.1 & 32.6 & 0.0 & 71.3 \\ 
        BAF-LAC~\cite{BAF-Net}           & 91.5 & 54.1 & 84.4 & 98.4 & 94.1 & 57.2 & 27.6 & 42.5 & 15.0 & 51.6 & 39.5 & 78.1 & 40.1 & 0.0 & 75.2 \\
        BAAF-Net~\cite{BAAFNet}          & 92.0 & 57.3 & 84.2 & 98.3 & 94.0 & 55.2 & 48.9 & 57.7 & 20.0 & 57.3 & 39.3 & 79.3 & 40.7 & 0.0 & 70.1 \\ 
        LGS-Net~\cite{LGSNet}            &  93.3 & \nd 63.6 & 86.1 & \rd 98.7 & 95.7 & 65.7 & \fs 62.8 & 52.6 & 36.5 & 62.0 & 52.1 & 84.3 & 45.9 & \nd 9.0 & 75.0 \\
        PushBoundary~\cite{PushBoundary} & \nd 93.8 & 59.7 & 85.8 & \fs 98.9 & \fs 96.8 & \fs 79.3 & 49.7 & 52.4 & 0.0  & 62.1 & \fs 57.6 & \fs 86.8 & 42.0 & 0.0 & 65.5 \\ 
        LCPFormer~\cite{lcpformer}   & 93.5 & \rd 63.4 & 86.5 & 98.3 & \rd 96.0 & 55.8 & 57.0 & 50.6 & \fs 46.3 & 61.4 & 51.5 & \rd 85.2 & \nd 49.2 & 0.0 & \nd 86.2 \\
        LACV-Net$^*$~\cite{LACV-Net}     & 93.2 & 61.3 & 85.5 & 98.4 & 95.6 & 61.9 & \rd 58.6 & \nd 64.0 & 28.5 & 62.8 & 45.4 & 81.9 & 42.4 & \rd 4.8 & 67.7 \\
        EyeNet~\cite{Yoo_2023_CVPRW} & \rd 93.7 &  62.3 & \rd 86.6 & 98.6 & \nd 96.2 & 65.8 & \nd 59.2 & \fs 64.8 & 17.9 & \nd 64.8 & 49.8 & 83.1 & \rd 46.2 & \fs 11.1 & 65.4 \\
        U-Next$^*$~\cite{zeng2023small}  & 93.0 &  62.8 & 85.2 & 98.6 & 95.0 & 68.2 & 53.6 &  60.4 & \rd 36.8 & \rd 64.0 & 48.9 & 84.9 & 45.1 & 0.0 & 76.2 \\
        
        \Ours{} (Ours)                   & \fs 94.0  & \fs 65.2 & \nd 86.7   & 98.3 & 95.8 & \nd 75.2 & 49.7   & \rd 60.5    & \nd 42.6 & \fs 66.3        & \rd 52.6   &  85.1 & \fs 50.9     & 1.2  & \rd 82.6  \\ \hline
        \end{tabular}
    }
    \vspace{\tabcapspace}
    \caption{\textbf{Comparison with SOTA methods on SensatUrban online benchmark~\cite{sensaturban_cvpr}}.
    Our method performs often better on rare classes which are difficult to label in one or the other domain.
    ${^*}$ indicates arXiv paper. Best results are highlighted as \colorbox{colorFst}{\bf first}, \colorbox{colorSnd}{second}, and \colorbox{colorTrd}{third}.}
    \label{tab:benchmark}
\vspace{-3mm}
\end{table*}
\subsection{Experimental setup} \label{sec:exp_setup}
\boldparagraph{SensatUrban Dataset.} SensatUrban~\cite{sensaturban_cvpr} is an urban-level photogrammetric point cloud dataset collected by a UAV. 
It covers a total of 7.64 square kilometers in three UK cities, \ie Birmingham, Cambridge and York, and provides annotations for 13 semantic categories.
Its average density of it is 473 points per square meter.
For easier processing, the data are cut into 43 blocks with a maximum size of 400 meters by 400 meters. 
We follow the official split, which consists of training/validation/testing set with 33/4/6 blocks.
During training and evaluation, the data from different cities are exploited mutually.
We use the training set for training and report ablation studies on the validation set.
We also report results on the testing set by submitting the predictions to the leaderboard where the ground truths are unpublished for a fair comparison. %
The grid size for down-sampling is set as 0.2 meters, resulting in 92\% of the original points being filtered out.
The pixel size for projection is set as 0.04 meters and the image size is set as $512 \times 512$, resulting in a coverage of $20.48m \times 20.48m$.

\boldparagraph{Metrics.}
As official recommendations~\cite{sensaturban_cvpr}, the main metric for per-category evaluation is intersection-over-union (IoU) and its mean value (mIoU) over all classes.
The IoU is formulated as follows:
\begin{equation}
    \text{IoU}_c =  \frac{{TP}_c}{{TP}_c+{FP}_c+{FN}_c} \enspace,
\end{equation}
where ${TP}_c$, ${FP}_c$ and ${FN}_c$ indicate true positive, false positive and false negative predictions for class $c$.
The mIoU is the average IoU over all classes:
\begin{equation}
    \text{mIoU} = \frac{1}{N_\text{cla}} \sum_{c=1}^{N_\text{cla}} \text{IoU}_c
\end{equation}
where ${N_\text{cla}}$ stands for the number of classes. 
Additionally, the overall accuracy is also reported. It is defined as follows:
\begin{equation}
    OA = \frac{\sum_{c=1}^{N_\text{cla}} {TP}_c}{N_\text{poitns}} \enspace,
\end{equation}
where $N_\text{points}$ is the total number of points.

\boldparagraph{Implementation details.}
HRNet~\cite{HRNet_base} with object contextual representation~\cite{OCRNet} and a variant of RandLA-Net~\cite{randlanet_2020} are chosen as the backbones for A-branch and P-branch, respectively.
These are detailed in the supplementary materials.
AdamW optimizer~\cite{AdamW} is used with a weight decay of 0.01 and a default learning rate of 0.001, while the learning rate of the P-branch is multiplied by a factor of 5.
The learning rate decreases by 5\% after each epoch.
The network is trained for 200 epochs for SensatUrban, with a batch size of 32.
During the training procedure, random rotation along z-axis, random flip along y-axis and random scale are performed for both grid-downsampled points and aerial images while the correspondences are preserved.
For more efficient training, the data in the training set and validation set are cropped into 100m $\times$ 100m patches approximately.

\subsection{Comparison with existing methods} \label{sec:sota_res}

\boldparagraph{Quantitative results.} 
The comparison of our method and other existing methods on SensatUrban benchmark~\cite{sensaturban_cvpr} are shown in Table~\ref{tab:benchmark}. 
Remarkably, \Ours{} surpasses all other methods, achieving an OA of 94.0\% and a mIoU of 65.2\%.
Notably, \Ours{} outperforms its backbone, RandLa-Net~\cite{randlanet_2020}, by an impressive margin of 12.5\%, affirming the beneficial impact of the A-branch on segmentation.
Furthermore, \Ours{} excels in specific categories, ranking first in both the traffic road and the footpath categories.
Additionally, \Ours{} attains a top-three position in 8 out of 13 categories, further validating its superior performance.

\begin{figure*}[t] \centering
    \raisebox{-0.5\height}{\includegraphics[width=0.48\textwidth]{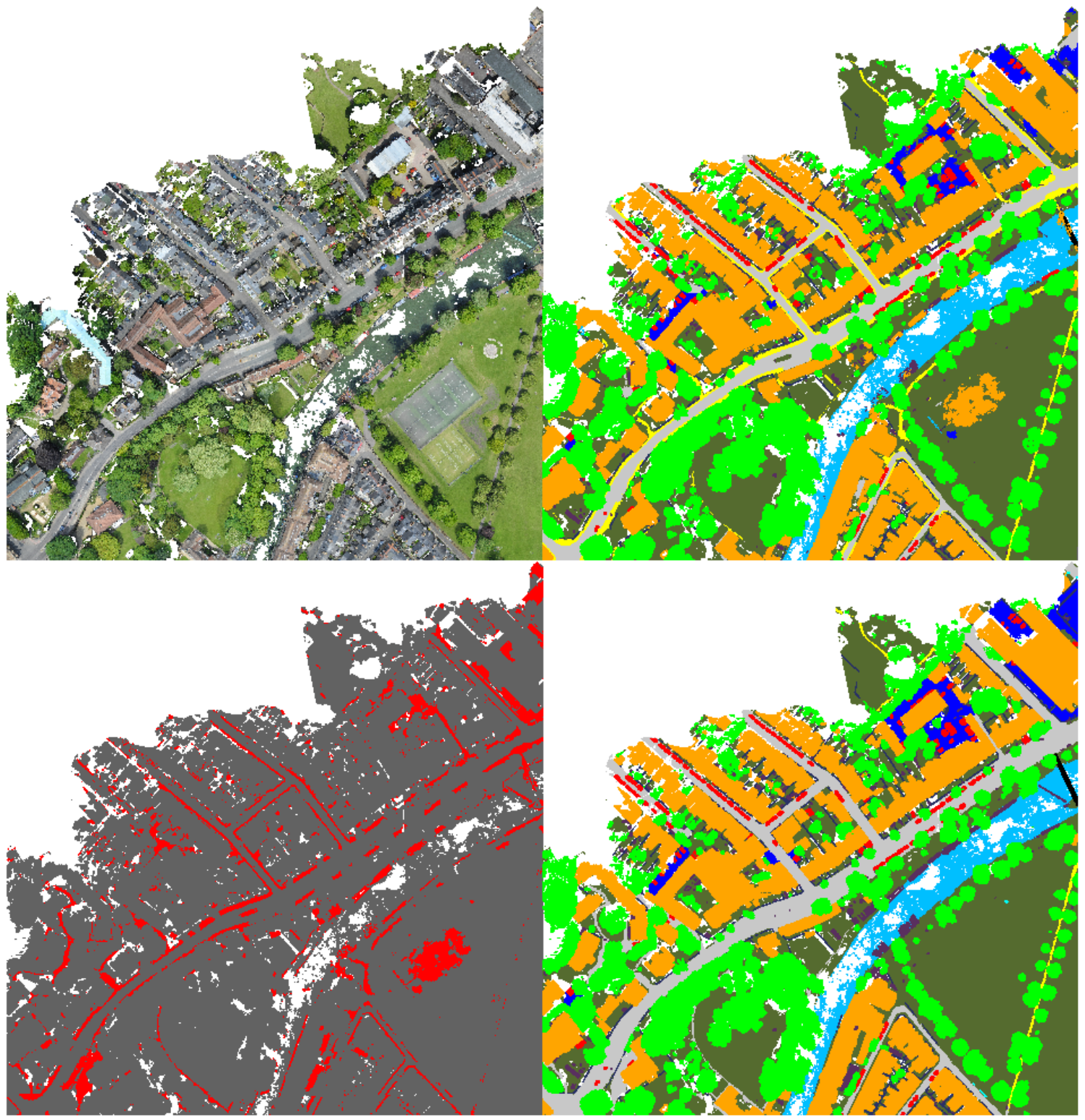}}
    \raisebox{-0.5\height}{\includegraphics[width=0.48\textwidth]{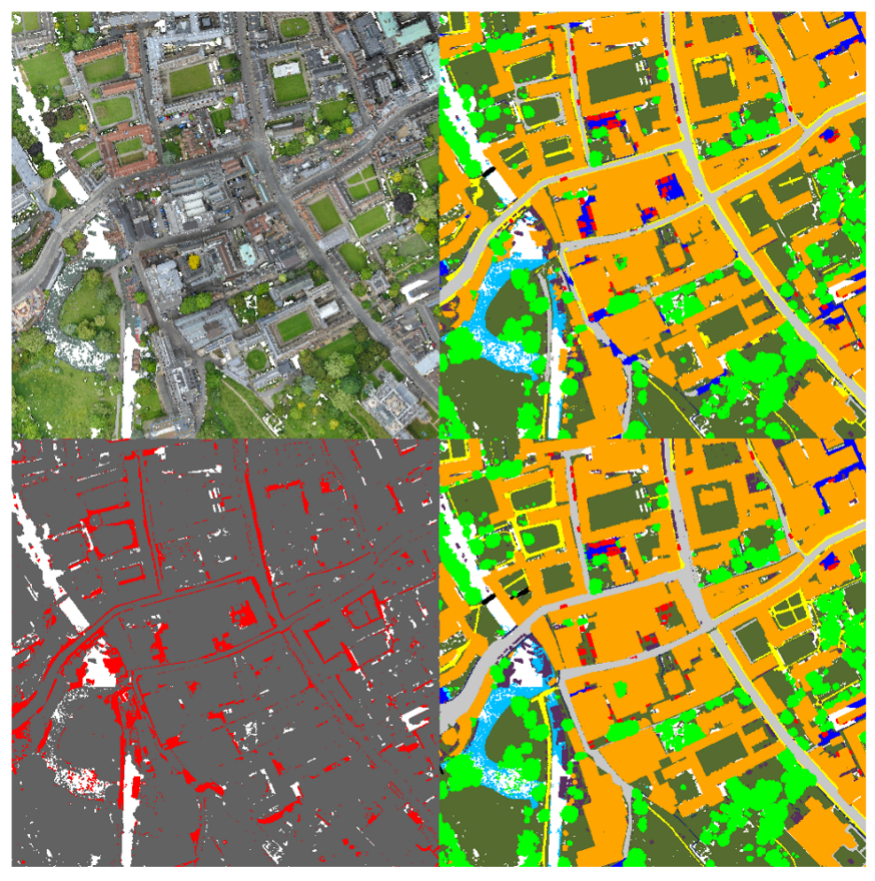}}
    \\ \vspace{0.2em}
    \includegraphics[width=0.96\linewidth]{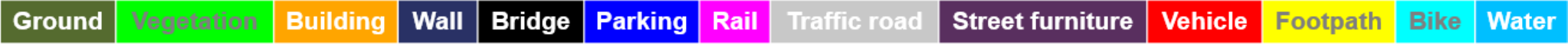}
    \\ \vspace{0.5em}
    \caption{\textbf{The qualitative result of two blocks in the validation set of SensatUrban~\cite{sensaturban_ijcv}.} In each figure, the top-left sub-figure is the aerial visualization of the original point cloud which covers a $400m \times 400m$ area. The top-right and bottom-right sub-figures are the predictions and ground truth respectively. Both of them follow the color bar at the bottom. The bottom-left sub-figure is an error map that presents the difference between the prediction and the ground truth.}
   \label{fig:qualitive_result}
\vspace{-3mm}
\end{figure*}

\begin{figure*}[t] \centering
    \makebox[0.29\textwidth]{\footnotesize Point cloud}
    \makebox[0.29\textwidth]{\footnotesize PushBoundary~\cite{PushBoundary}}
    \makebox[0.29\textwidth]{\footnotesize \Ours{} (Ours)}
    \\
    \includegraphics[width=0.29\textwidth]{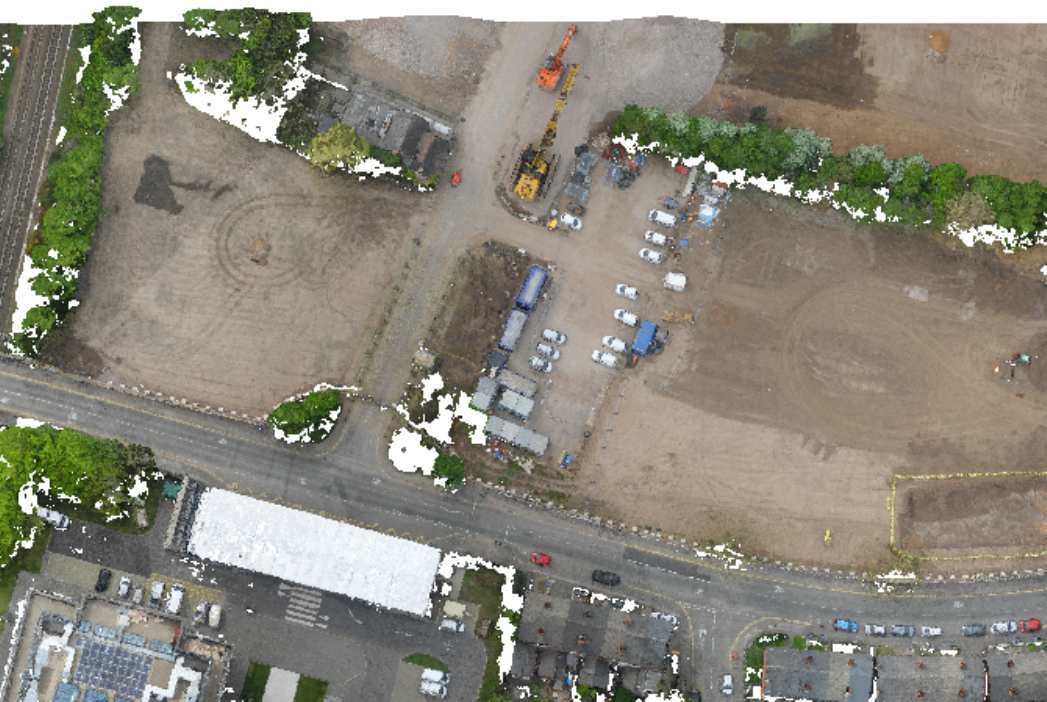}
    \includegraphics[width=0.29\textwidth]{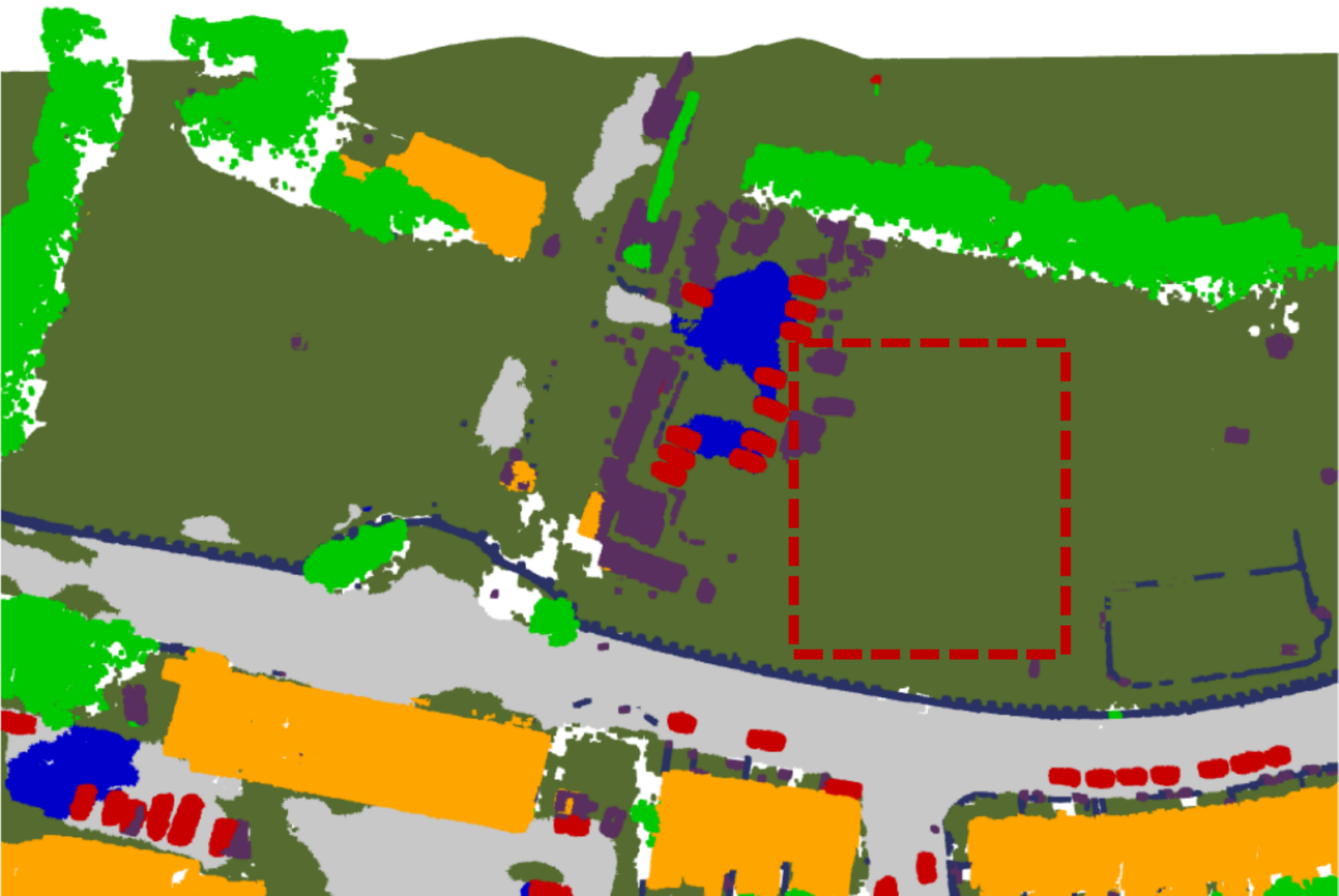}
    \includegraphics[width=0.29\textwidth]{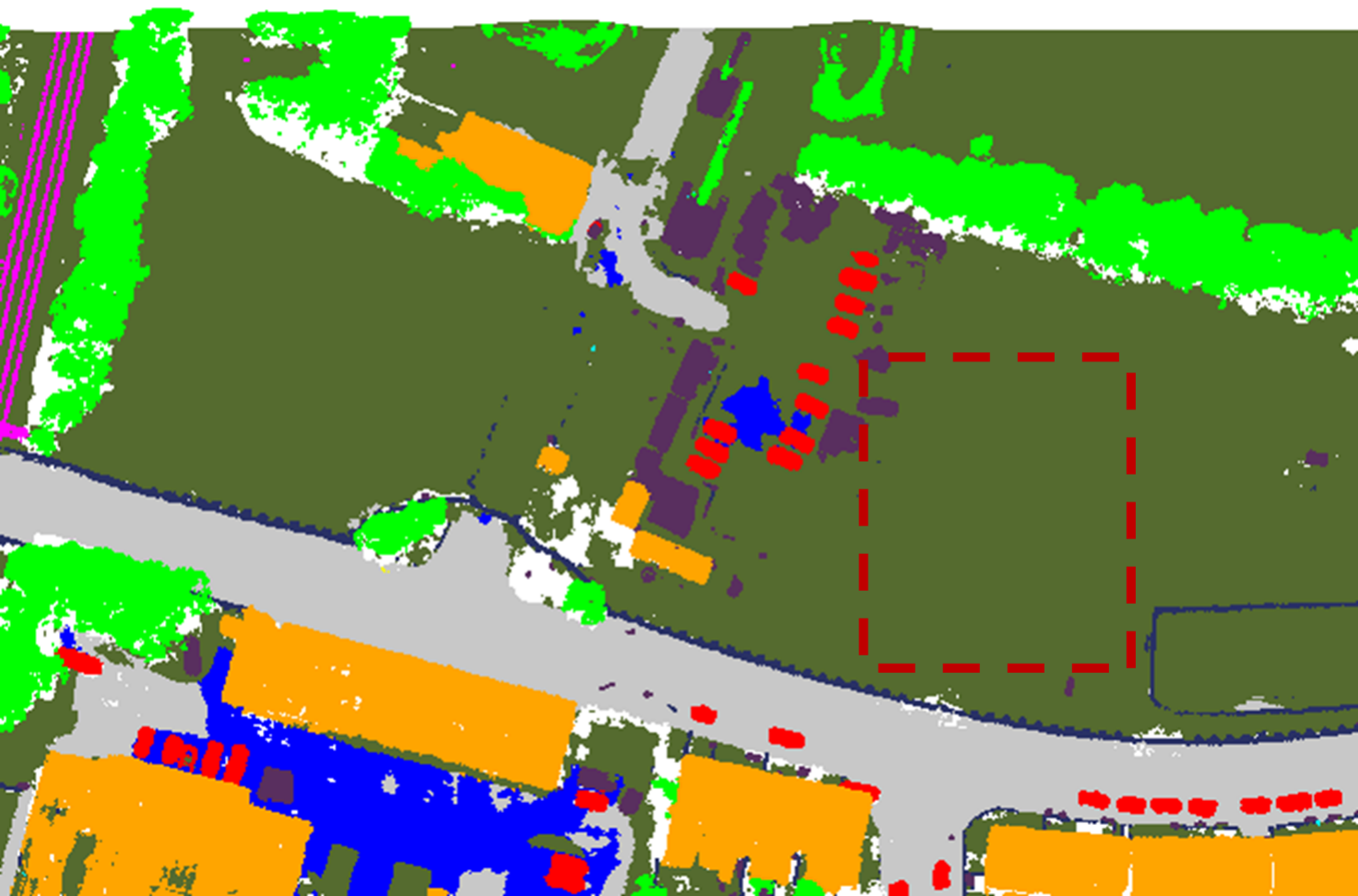}
    \\
    \includegraphics[width=0.29\textwidth]{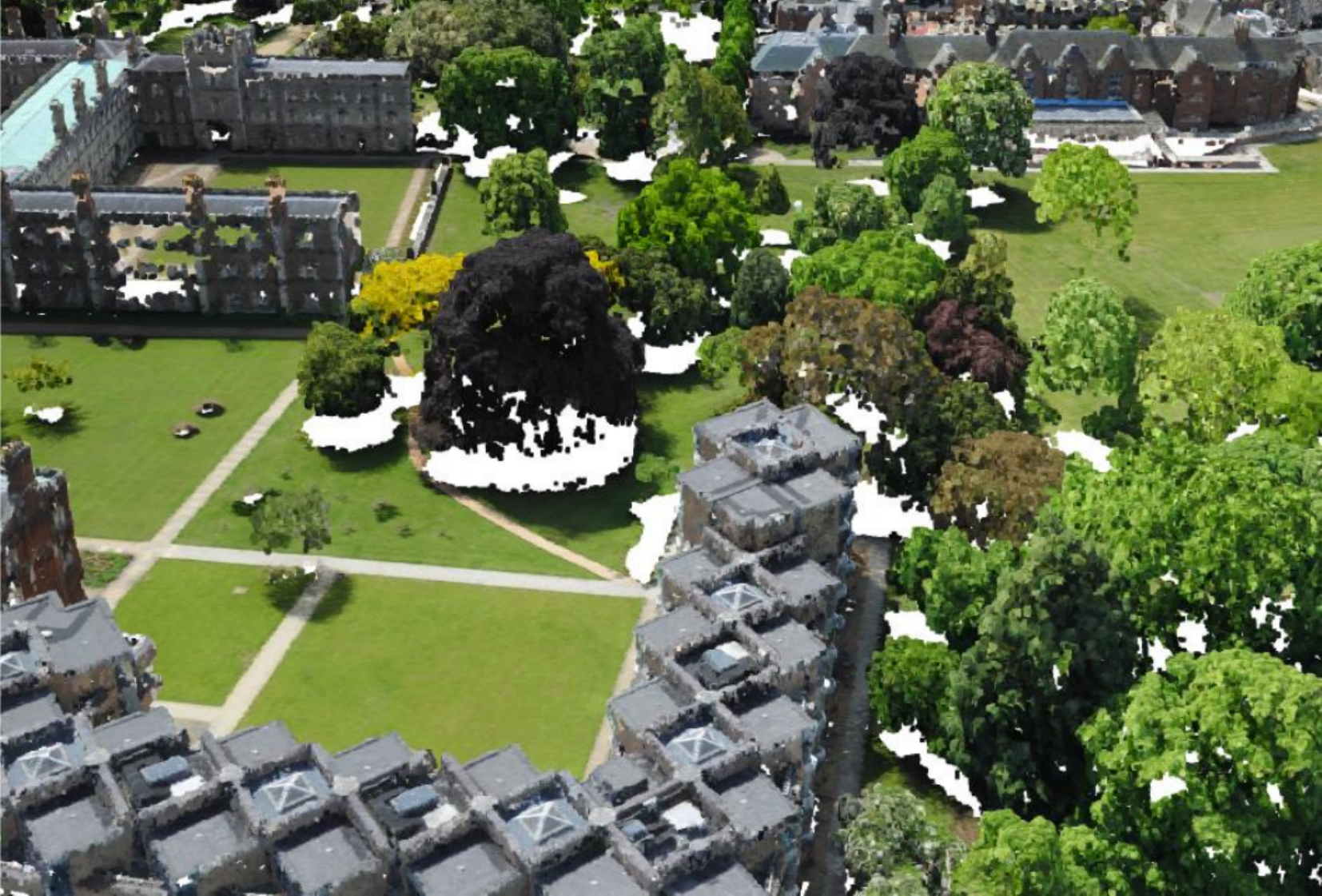}
    \includegraphics[width=0.29\textwidth]{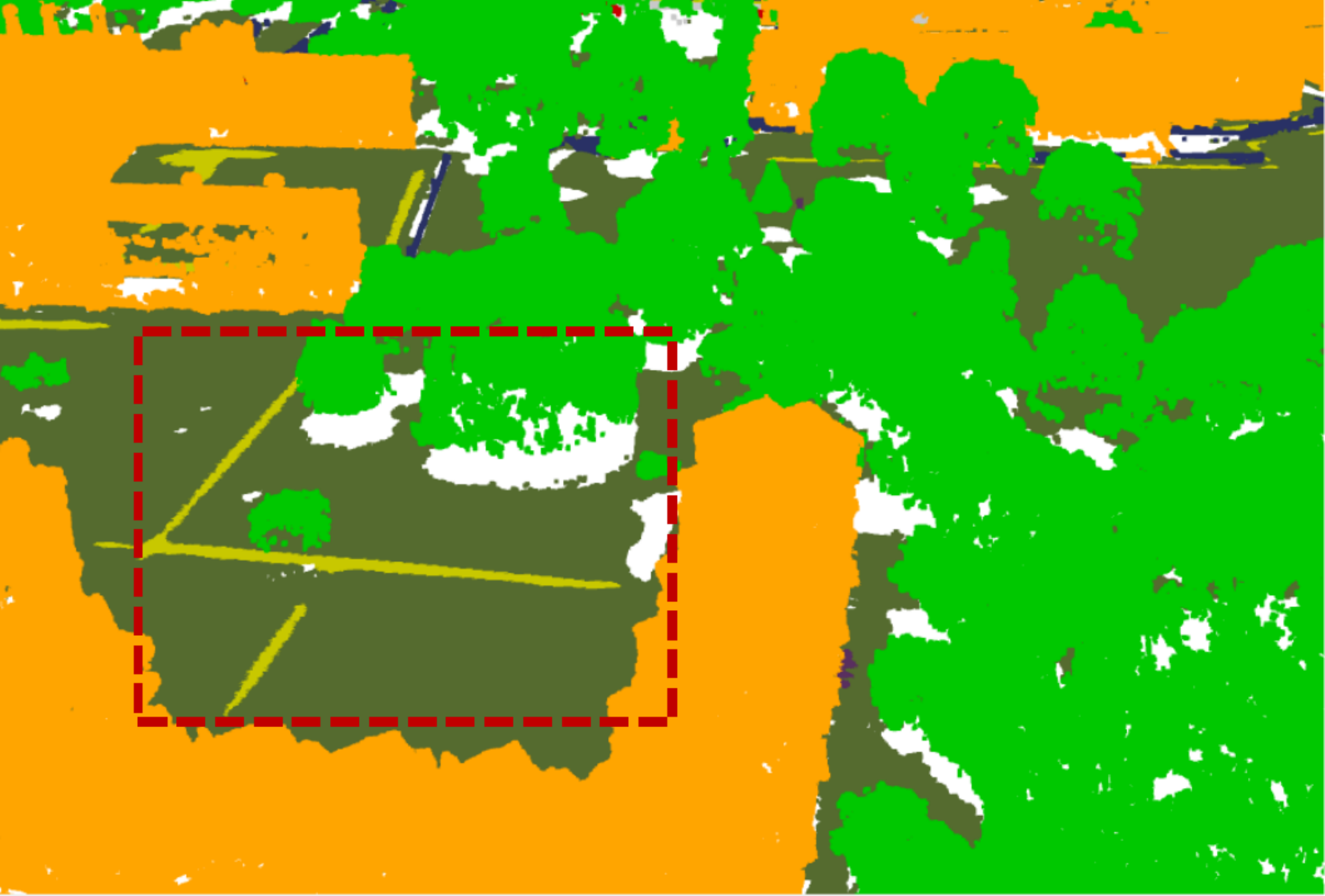}
    \includegraphics[width=0.29\textwidth]{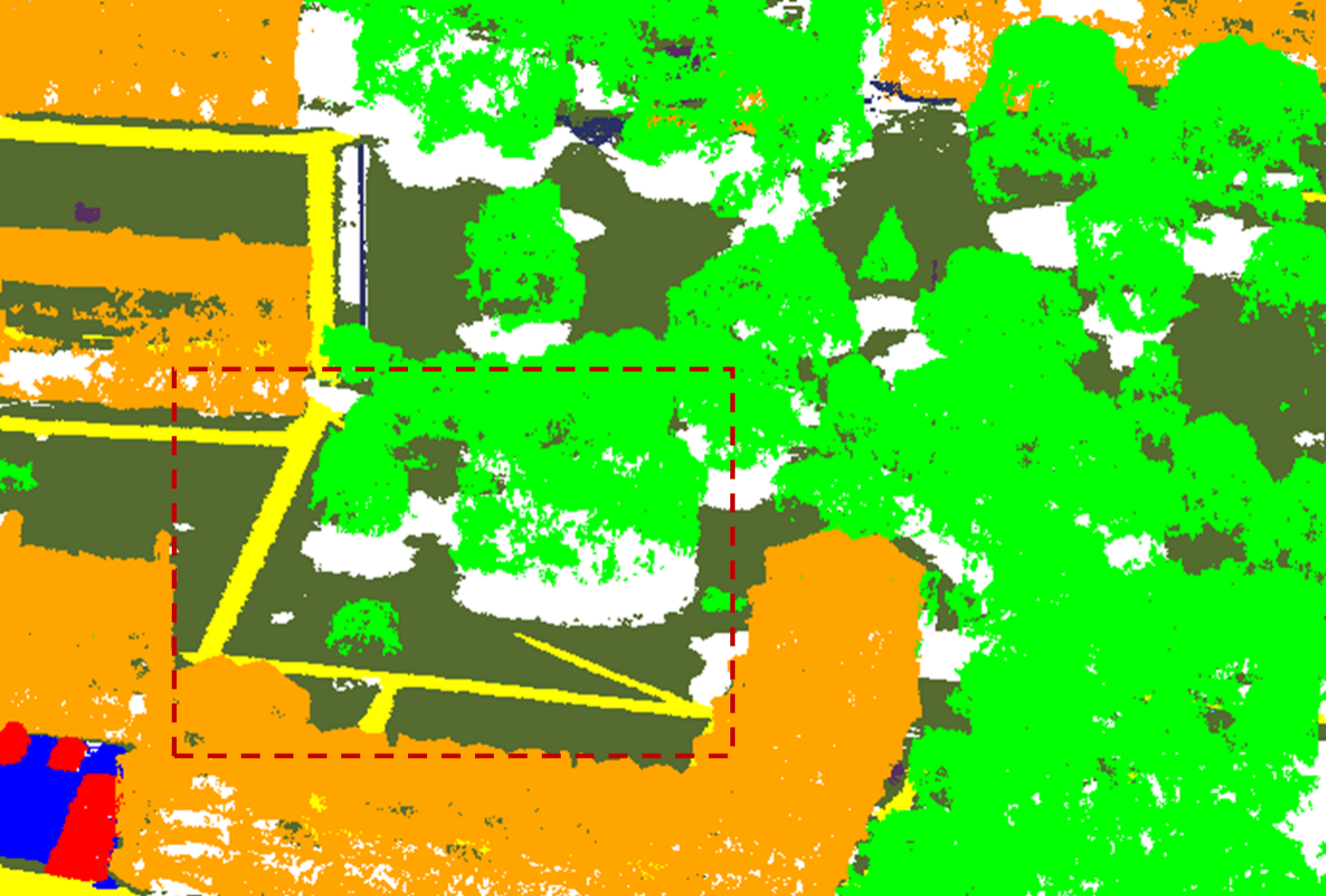}
    \\ \vspace{0.2em}
    \includegraphics[width=0.89\linewidth]{figures/QualitativeDetailsSensat/Sensat_Color_Bar_double-column.pdf}
    \\ \vspace{0.2em}
    \caption{\textbf{Qualitative comparison with PushBoundary~\cite{PushBoundary} on the SensatUrban~\cite{sensaturban_cvpr} test set (No GT available).} The figures of the PushBoundary with the red dash boxes are directly taken from the original paper. \Ours{} performs on par with PushBoundary in the first example (top row) and outperforms it in the second example (bottom row).}
    \label{fig:comp_pushboundary}
\vspace{-1mm}
\end{figure*}

\boldparagraph{Qualitative results.}
Fig.~\ref{fig:qualitive_result} is a high-level visualization to qualitatively compare the prediction of \Ours{} and the ground truth.
As indicated by the OA, \Ours{} predicts most of the points correctly and performs excellently in the two $400m \times 400m$ blocks.
Nevertheless, the primary source of inaccuracy in this figure is from the footpath, which presents problems due to its contextual and physical resemblance to the traffic road.
Fig.~\ref{fig:comp_pushboundary} showcases a visual assessment of \Ours{} against PushBoundary~\cite{PushBoundary}.
The middle column, \ie the results of PushBoundary with the red dashed boxes, is taken directly from the original paper. Even though the target regions are chosen by other authors, our method shows comparable or superior performance compared to PushBoundary.

\subsection{Ablation studies} \label{sec:ablation}
\newcommand{\bd}{\bf}   %

\begin{table*}[t] \centering    
    \resizebox{\textwidth}{!}{
        \begin{tabular}{lccccccccccccccc}
        \hline
          Method &
          OA &
          mIoU &
          \spheading{ground} & 
          \spheading{vegetation} & 
          \spheading{building} & 
          \spheading{wall} & 
          \spheading{bridge} & 
          \spheading{parking} & 
          \spheading{rail} & 
          \spheading{traffic road} & 
          \spheading{street furniture} & 
          \spheading{car} & 
          \spheading{footpath} & 
          \spheading{bike} & 
          \spheading{water} \\ 
        \hline
        
        A-branch      & \rd 89.8 & \nd 55.2 & \rd 71.8 & \rd 91.6 & \nd 94.3 & \nd 70.0 & \nd 22.9 & \nd 47.2 & \fs 46.6 & \nd 65.4 & \nd 45.4 & \nd 81.1 & \fs 20.4 & \rd 0.0 & \rd 61.0 \\ 
        P-branch      & \nd 90.2 & \rd 52.1 & \nd 75.0 & \nd 95.4  & \rd 93.3 & \rd 52.4 & \fs 27.4 & \rd 40.7 & \rd 23.3 & \rd 59.3 & \rd 34.3 & \rd 80.6 & \rd 18.2 & \fs 12.4  & \nd 65.2 \\
        \Ours{} (Ours) & \fs 92.3 & \fs 59.2 & \fs  80.5 & \fs 97.4 & \fs 96.7 & \fs 73.0 & \rd 21.8 & \fs 52.3 & \nd 43.4 & \fs 66.1 & \fs 50.7 & \fs 84.8 & \nd 19.9 & \nd 12.3 & \fs 70.9 \\

        \hline
        \end{tabular}
    }
    \vspace{\tabcapspace}
    \caption{\textbf{Ablation studies on branches.} This table compares the semantic labeling performance of the aerial image branch and the point cloud branch against the output of the geometry-aware fusion module. The benefit of the fusion module is apparent as is mostly yields better class-wise performances then the individual branches separately. }
    \label{tab:branches}
\vspace{-3mm}
\end{table*}
\boldparagraph{Branch ablations.}
We first compare A-branch, P-branch and \Ours{}.
For the single branch networks, the \OursFusion~strategy is not applied as the features are obtained from single representation. 
The output feature from A/P-branch is directly passed to the segmentation head and generates an intermediate prediction.
For A-branch, the final prediction is generated through a bilinear interpolation based on four neighbouring pixels.
For P-branch, the final prediction is obtained by coping prediction from the nearest neighbour point within downsampled point set.
As shown in Table~\ref{tab:branches}, the combined network outperforms every single branch on OA, mIoU and most of the IoUs of all categories.
In cases where \Ours{} performs worse than single-branch networks, the difference is negligible.
Notably, P-branch outperforms A-branch on OA, although the opposite is observed for most categories.
This is because of the imbalanced distribution of categories in the dataset.
Over 50\% of the points are attributed to the three categories - ground, vegetation, and building, resulting in that a higher accuracy for these dominant categories will mask shortcomings in other categories for an overall metric.

\begin{table}[t]  
    \centering
    \setlength{\tabcolsep}{10pt}
    \begin{tabular}{llcc}
    \hline
    Encoder                         & Fusion strategy & OA   & mIoU \\ \hline
    A-branch                        & N/A             & 89.8 & 55.2 \\
    P-branch                        & N/A             & 90.2 & 52.1 \\ \cline{1-1}
    \multirow{4}{*}{Dual-encoder}   & Addition         & \rd 91.3 & \rd 56.7 \\
                                    & Concatenation   & 90.7 & \rd 56.7 \\
                                    & Naive GAF       & \nd 91.5 & \nd 57.5 \\
                                    & GAF             & \fs 92.3 & \fs 59.2 \\ \hline
    \end{tabular}
    \vspace{\tabcapspace}
     \caption{\textbf{Ablation studies on geometry-aware fusion (GAF) module.} %
     Compared to the simpler point-wise fusion approaches (addition, concatenation), the geometry-aware fusion includes spatial context into the reasoning yielding improved performance.
     }
    \label{tab:fusion}
\vspace{-3mm}
\end{table}

\boldparagraph{Fusion strategy.}
We compare \OursFusion~module with two simple fusion strategies and the naive version of \OursFusion~in Table~\ref{tab:fusion}.
The addition is the most intuitive way to combine two features.
The concatenation increases the complexity slightly because a subsequent MLP is necessary to reduce the number of channels.
These two combinations are point-wise and thus no neighbouring features are considered.
Nevertheless, they outperform both single-branch networks.
Naive GAF enhance its local adaptive capabilities by involving neighbour features at a downsampled points level.
The proposed GAF improves the naive GAF by using the original points as query points and achieves the best performance on both OA and mIoU. Our GAF module yields enhanced outcomes, surpassing the simple fusion strategy by 1\% OA and 2.5\% mIoU, \eg addition and concatenation. Ablation studies illustrate the effectiveness and necessity of each component in the proposed method.

\section{Conclusion} \label{sec:conclusion}
We presented a semantic segmentation method that exploits the advantages of both point cloud-based and aerial image-based methods in a single network architecture with two separate domain branches.
The reasoning about which branch is more effective for which class category and spatial location is learned by a geometry-aware fusion network that combines the output of both branches into a single estimate.
Ablation studies and comparisons to state-of-the-art methods show clear benefits of the proposed architecture.

\section{Acknowledgements}
This work is financially supported by TomTom, the University of Amsterdam and the allowance of Top consortia for Knowledge and Innovation (TKIs) from the Netherlands Ministry of Economic Affairs and Climate Policy. Fatemeh Karimi Nejadasl is financed by the University of Amsterdam Data Science Centre.

\appendix

\section{Implementation details} \label{sec:supp_details}
\boldparagraph{A-branch.} We adopt HRNet~\cite{HRNet_base} with object-contextual representations (OCR)~\cite{OCRNet}, denoted as HRNet-OCR, as the backbone for A-branch.
During training, the OCR loss is preserved while the original 2D segmentation head is removed.
The intermediate features, also known as augmented representations as defined in the original paper, from HRNet-OCR are compressed to a total of 128 channels, thereby ensuring alignment with the output of the P-branch.

\boldparagraph{P-branch.} We employ RandLA-Net~\cite{randlanet_2020} as the backbone for the P-branch and follow its official configuration for the SemanticKITTI dataset\cite{SemanticKITTI} with the following two modifications:
Firstly, we double all feature channels in the RandLA-Net to accommodate the additional color features.
Furthermore, we double the output channel for the last layer to ensure compatibility with the A-branch.
Consequently, the encoder produces outputs with channel dimensions of 64, 128, 256, and 512, respectively.
Secondly, we input the same point cloud to RandLA-Net twice and sum up the output features. %
Although the network does not change, the down-sampling within the network is random, leading to different features for the same point cloud in the end.
This technique promotes the consistency of RandLA-Net.

\boldparagraph{GAF module.}
We adopt KPConv~\cite{thomas_kpconv_2019} as the point convolution in the GAF module and adhere to the configuration of the rigid KPConv.
Accordingly, one single rigid KPConv encompasses a sphere with a radius of 0.5 meters, centered at the query point.
Each kernel point exerts an influence on all support points within a sphere whose radius is 0.24 meters and centered on the kernel point.

{\small
\bibliographystyle{ieee_fullname}
\bibliography{egbib}
}

\end{document}